\pdfoutput=1

\documentclass[11pt]{article}

\usepackage[]{acl}

\usepackage{times}
\usepackage{latexsym}

\usepackage[T1]{fontenc}

\usepackage[utf8]{inputenc}

\usepackage{microtype}
\usepackage{graphicx}
\usepackage{amsmath}
\usepackage{amssymb}
\usepackage{multirow}
\title{Value Retrieval with Arbitrary Queries for Form-like Documents}

\author{\textbf{Mingfei Gao}\thanks{\ \ Mingfei and Le contributed equally.} \ , \textbf{Le Xue}$^{*}$, \textbf{Chetan Ramaiah}, \textbf{Chen Xing}, \textbf{Ran Xu}, \textbf{Caiming Xiong} \\
Salesforce Research, Palo Alto, USA \\
{\tt\small{\{mingfei.gao, lxue, cramaiah, cxing, ran.xu, cxiong\}@salesforce.com}}
  }

\begin{document}
\maketitle
\begin{abstract}
We propose value retrieval with arbitrary queries for form-like documents to reduce human effort of processing forms. Unlike previous methods that only address a fixed set of field items, our method predicts target value for an arbitrary query based on the understanding of the layout and semantics of a form. To further boost model performance, we propose a simple document language modeling (SimpleDLM) strategy to improve document understanding on large-scale model pre-training. Experimental results show that our method outperforms previous designs significantly and the SimpleDLM further improves our performance on value retrieval by around 17\% F1 score compared with the state-of-the-art pre-training method.  \href{https://github.com/salesforce/QVR-SimpleDLM}{\emph{Code is available here}}.
\end{abstract}

\section{Introduction}

\begin{figure}[t]
    \centering
    \includegraphics[width=1.0\linewidth]{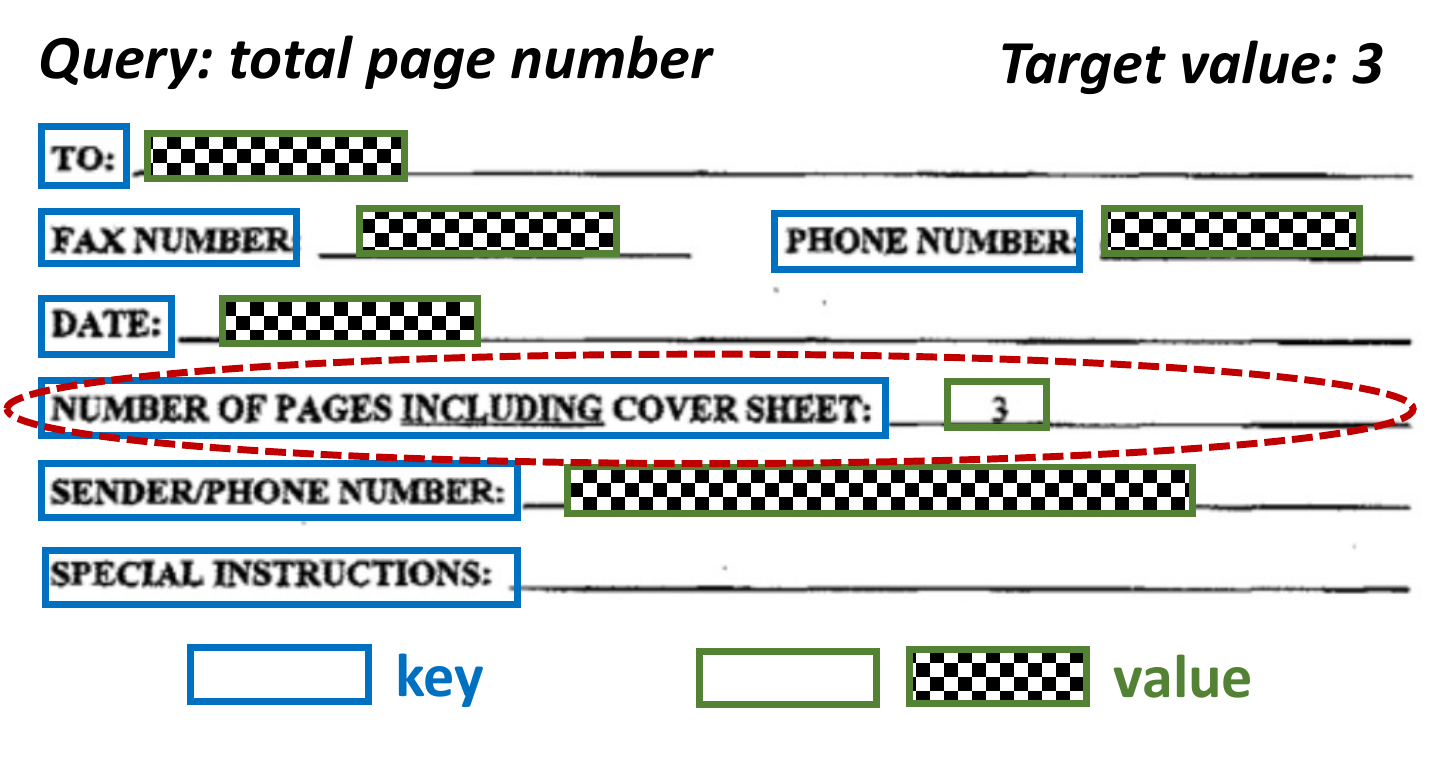}
    \caption{Illustration of the value retrieval with an arbitrary query. A user obtains the target value from a document based on a query of interest. Some values are decorated with a mosaic for privacy purposes.}
    \label{fig:form_example}
     \vspace{-5mm}
\end{figure}

Form-like documents are very commonly used in business workflows. However, tremendous forms are still processed manually everyday. When humans need to extract some relevant information from a form-like document, they proceed as conducting a value retrieval task with some text queries. For example in Figure~\ref{fig:form_example}, when humans process a form, they usually have a description of the information (query) that they want to extract (e.g., total page number). Then, they examine the form (usually an image or a PDF) carefully to locate the key (e.g., \emph{NUMBER OF PAGES INCLUDING COVER SHEET} in Figure~\ref{fig:form_example}) that is most semantically similar to the query and finally infer the target value based on the localized key. This manual process costs a large amount of human efforts as the number of forms and queries increases. Automating information extraction from forms is important to alleviate this problem.

Existing methods formulate the problem as sequence labeling~\cite{xu2020layoutlm} or field extraction~\cite{gao2021field}, where they define a fixed set of items of interest (referred as fields) and train models that only extract values of the pre-defined fields. There are at least two limitations of this formulation. First, forms are very diverse and it is impossible to cover all the items of interest using a fixed set of fields. Second, their models are very domain-specific and hard to be utilized for different form types. For example, an invoice field extractor may not be able to process resumes, since different fields are expected for these two form types.

To handle diverse queries with a unified model, we formulate the problem as value retrieval with arbitrary queries for form-like documents. 
Under such task formulation, users can extract values from a form by presenting variants of the corresponding keys as queries.
We also set up a benchmark for the task by introducing a simple yet effective method. 
The method takes an arbitrary query phrase and all the detected optical character recognition (OCR) words with their locations in the form as inputs. 
Then, we model the interactions between the query and the detected words from the document using a transformer-based architecture. The training objective encourages the matching of the positive query-value pairs and discourages that of the negative ones. To further boost the performance, we present SimpleDLM, a simple document pre-training strategy that makes it more flexible to learn local geometric relations between words/tokens compared to existing pre-trained models. Experimental results show that our method outperforms the baselines by a large margin under different settings. When initializing using SimpleDLM, our method is further improved largely by about 17\% F1 score compared to initializing using a state-of-the-art (SOTA) pre-trained model, i.e., LayoutLM~\cite{xu2020layoutlm}.

\section{Related Work}
\label{sec:related_work}
\noindent\textbf{Information Extraction from Documents} is crucial for improving the efficiency of form processing and reducing human labor. Information extraction is often formulated as a field extraction task. \citealt{palm2019attend} propose an invoice field extractor by using an Attend, Copy, Parse architecture. ~\citealt{majumder2020representation} present a field-value pairing framework that learns the representations of fields and value candidates in the same feature space using metric learning. \citealt{nguyen2021span} propose a span extraction approach to extract the start and end of a value for each field. \citealt{gao2021field} introduce a field extraction system that can be trained with large-scale unlabeled documents. \citealt{xue2021robustness} propose form transformations to mimic the variations of forms for robustness evaluation. Unlike previous methods that aim to extracting values for a pre-defined set of fields, our method targets at retrieving values for arbitrary queries. 

\noindent\textbf{Document Pre-training} is an effective strategy to improve document-related downstream tasks. \citealt{xu2020layoutlm} propose LayoutLM that models interaction between texts and layouts in scanned documents using masked language modeling and image-text matching. Later, LayoutLMv2~\cite{xu2020layoutlmv2} introduces a spatial-aware self-attention mechanism to improve learning relative positional relationship among different text blocks. Most recently, \citealt{appalaraju2021docformer} propose DocFormer that encourages the interaction between image and text modalities by adding an image reconstruction task. The existing pre-training methods perform well when applied to downstream tasks such as document classification and token sequence labeling. However, all of the above methods include the absolute 1-D positional embedding (the so called reading order) of the tokens in the inputs. Although this 1-D embedding is a helpful prior knowledge for a holistic understanding of a document, it hinders a model from learning rich geometric relationship among tokens, thus is not beneficial to our value retrieval task, where local geometric relationship between words is essential for prediction. Our method uses permutation-invariant positional encoding to improve model performance.

\begin{figure*}[t]
    \centering
    \includegraphics[width=1.0\linewidth]{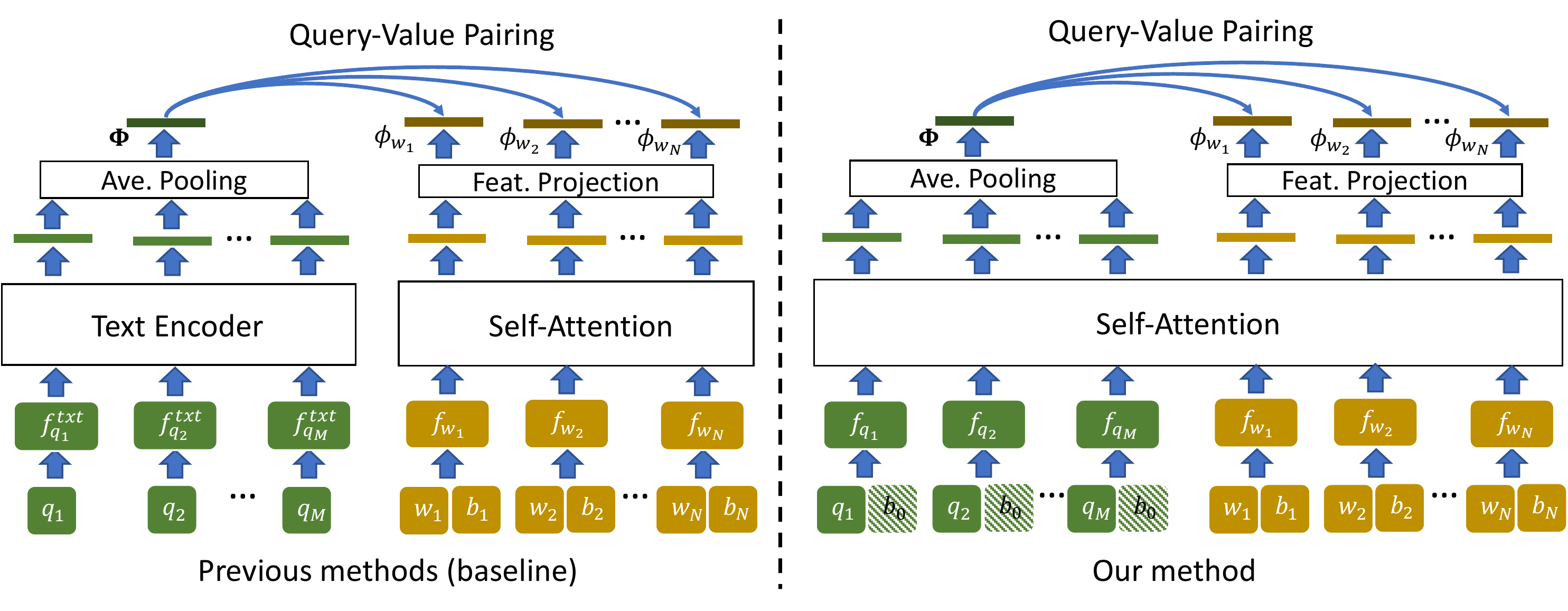}
    \vspace{-8mm}
    \caption{Comparison of previous methods and our method (see details in Section~\ref{sec:approach}).}
    \label{fig:method}
    \vspace{-6mm}
\end{figure*}
\section{Our Approach}
\label{sec:approach}
\noindent\textbf{Problem Formulation}. The inputs to the system are an expected key phrase as the query, $\mathbf{Q}_i$, and a document $\mathbf{D}_i$. $\mathbf{D}_i$ is represented using a set of OCR words $\{w_{i1},w_{i2},...,w_{iN}\}$ and their bounding-box locations $\{b_{i1},b_{i2},...,b_{iN}\}$ in the document. The input query phrase is tokenized to $\mathbf{Q}_i=\{q_{i1}, q_{i2},...,q_{iM}\}$. A value retrieval system reads the document and understands the layout and semantics. The goal is to pick a phrase from the OCR words as the value $\mathbf{V}_i$ for the input query, $\mathbf{Q}_i$. Since the modeling is within one document, we will omit the subscript $i$ for simplicity.

\subsection{Value Retrieval with Arbitrary Queries}
To the best of our knowledge, there are no existing methods explicitly address value retrieval with arbitrary queries from scanned forms. However, it is possible to apply the recent pairing-based field extraction methods~\cite{majumder2020representation,nguyen2021span} to this task with some modification. In previous design, they first embed fields using a text encoder, extract OCR word representation by modeling the interaction of OCR words using self-attention and then conduct the field-value pairing. To accommodate arbitrary queries, we can simply replace their field (in a fixed set) embedding with the embedding of an arbitrary query as shown in Figure~\ref{fig:method} (previous method). There are two obvious drawbacks of these methods: (1) they model the interaction between query and OCR words in a shallow way using only simple fully connected layers and (2) they require an additional model of text encoder for query embedding which introduces extra computational cost. We set this previous design as our baseline. The implementation details are shown in the appendix.

In contrast, our method utilizes a unified model to deeply model interactions among the query words and the OCR words. The direct inputs are a query $\{q_{1}, q_{2},...,q_{M}\}$ and OCR words $\{w_{1},w_{2},...,w_{N}\}$ associated with their locations $\{b_{1},b_{2},...,b_{N}\}$. We use a fixed dummy location $b_0$ for each query. Each query/OCR word is embedded as $f^{txt}_{w_{j}}\in \mathbb{R}^{d}$ and its location is encoded as $f^{loc}_{b_{j}}\in \mathbb{R}^{d}$, where $d$ indicates the length of each vector (see Section~\ref{sec:simpleDLM} for details). The final embedding of each word (e.g., $\mathbf{f}_{q_{j}}$ for a query word or $\mathbf{f}_{w_{j}}$ for an OCR word) is the summation of its word embedding and location embedding.

A transformer is used to model the interactions among $\{\mathbf{f}_{q_{1}},\mathbf{f}_{q_{2}}..,\mathbf{f}_{q_{M}}\}$ and $\{\mathbf{f}_{w_{1}},\mathbf{f}_{w_{2}},...,\mathbf{f}_{w_{N}}\}$ via $L$ self-attention layers. In the $l^{th}$ layer, the hidden representation of the $j^{th}$ token is updated following
\vspace{-5mm}
\begin{align}
    \mathbf{h}_j^l & =  \text{Softmax}(\frac{\mathbf{h}_j^{l-1}{\mathbf{H}^{l-1}}^T}{\sqrt{d_h}})\cdot\mathbf{H}^{l-1},
    \label{eq:self-attn}
\end{align}
\vspace{-1mm}
where $\mathbf{H}^{l-1}$ indicates the representation of all tokens from the $(l-1)^{th}$ layer and $d_h$ denotes the length of the hidden feature $\mathbf{h}_j^l$. $\mathbf{H}^{0}=\{\mathbf{f}_{q_{1}},\mathbf{f}_{q_{2}},...,\mathbf{f}_{q_{M}}, \mathbf{f}_{w_{1}}, \mathbf{f}_{w_{2}}...,\mathbf{f}_{w_{N}}\}$. In the final layer, $\mathbf{H}^{L}=\{\hat{\mathbf{f}}_{q_{1}},\hat{\mathbf{f}}_{q_{2}},...,\hat{\mathbf{f}}_{q_{M}}, \hat{\mathbf{f}}_{w_{1}}, \hat{\mathbf{f}}_{w_{2}}...,\hat{\mathbf{f}}_{w_{N}}\}$. The self-attention mechanism deeply models the interactions among query words and OCR words.

We obtain the query phrase representation, $\Phi$, using average pooling over $\{\hat{\mathbf{f}}_{q_{1}},\hat{\mathbf{f}}_{q_{2}},...,\hat{\mathbf{f}}_{q_{M}}\}$. The final representation of $w_j$ is obtained by $\phi_{w_{j}}=FC(\hat{\mathbf{f}}_{w_{j}})$, where $FC$ indicates a fully connected layer. The likelihood score of $w_j$ being a part of the target value of the query is obtained in Equation~\ref{eq:similarity_score}
\vspace{-2mm}
\begin{align}
    \mathbf{s}_j & =  \text{Sigmoid}(\phi_{w_{j}} \cdot \Phi^T).
    \label{eq:similarity_score}
\end{align}
\vspace{-1mm}
Our model is expected to learn (1) the layout and semantics of the document (2) the mapping between the input query phrase and the actual key texts in the document and (3) the geometric and semantic relationship between the key and value.

\noindent\textbf{Model optimization}. During training, each $w_j$ is associated with a ground-truth label $y_j\in\{0,1\}$, where $1$ means this word is a part of the target value and $0$ means it is not. The model is optimized using binary cross entropy loss as $\frac{1}{N}\sum_{j=1}^N y_j\text{log}\mathbf{s}_j+(1-y_j)(1-\text{log}\mathbf{s}_j)$.

\noindent\textbf{Inference}. Since the target value may contain multiple words, we group nearby OCR words horizontally as value candidates, $\{g_1, g_2, ..., g_D\}$, based on their locations using DBSCAN algorithm~\cite{ester1996density}, where $D$ is the number of grouped candidates. The value score of each candidate, $g_r$, is the maximum of all its covered words as $\mathbf{S}_r = \max_{w_j \in g_r} \mathbf{s}_j$, where, $w_j \in g_r$ indicates the OCR word $w_j$ is a part of the grouped $g_r$. The value candidate with the highest score is used as the value prediction.
\vspace{-2mm}

\subsection{SimpleDLM for Document Pre-training}
\label{sec:simpleDLM}
We introduce a Simple Document Language Modeling (SimpleDLM) method to encourage the understanding of the geometric relationship among words during pre-training.

The inputs to our pre-trained model are the OCR words $\{w_{1},w_{2},...,w_{N}\}$ associated with their locations $\{b_{1},b_{2},...,b_{N}\}$.  The word/location embedding protocol and the transformer structure of our pre-trained model are the same as those of our value retrieval model such that the pre-trained model can be directly used for initializing the parameters of the value retrieval model. Specifically, the word embedding, $f^{txt}_{w_{j}}$, is constructed by using a simple look up table. Previous works~\cite{xu2020layoutlm,xu2020layoutlmv2,appalaraju2021docformer} require the input text sorted in the reading order, so that they can process texts of the document in a similar way of processing languages. They leverage this prior knowledge by adding the ranking of each word as the 1-D positional embedding in the final location embedding, $f^{loc}_{b_{j}}$. This 1-D embedding provides a holistic view of the geometric relationship of words. However, it introduces extra dependence on the OCR engines (most SOTA OCR engines do not have the capability of sorting detected OCR words in the reading order), and it also restricts the model from learning local geometric relations between words in a flexible way. To encourage a model to better learn the local geometric relations, we exclude the 1-D positional embedding and only encode the 2-D bounding-box location (top-left, bottom-right, width and height) of each word using a lookup table. For simplicity, we only use the masked language modeling as the pre-training objective. We show in Section~\ref{sec:experiment} that our method improves the SOTA largely by using this simple pre-training strategy.

\section{Experiments}
\label{sec:experiment}
The following datasets are used in our experiments.

\noindent\textbf{IIT-CDIP}~\cite{lewis2006building} is a large-scale unlabeled document dataset that contains more than 11 million scanned images. Following prior works~\cite{xu2020layoutlm,xu2020layoutlmv2,appalaraju2021docformer}, our model is pre-trained using this dataset.

\noindent\textbf{FUNSD}~\cite{jaume2019funsd} is a commonly used dataset for spatial layout analysis. It contains 199 scanned forms with 9,707 semantic entities annotated, where 149 samples are for training and 50 for testing. The semantic linking annotations for all the key-value pairs are provided in the dataset.

\noindent\textbf{INV-CDIP}~\cite{gao2021field} is document dataset which contains 350 real invoices for testing. This dataset has key-value pair annotations for 7 commonly used invoice fields including \emph{invoice\_number}, \emph{purchase\_order}, \emph{invoice\_date}, \emph{due\_date}, \emph{amount\_due}, \emph{total\_amount} and \emph{total\_tax}. We evaluate our model using this test set.

\noindent\textbf{Settings}.
By default the annotated key texts of each dataset is used as the queries. The location of the keys are not used. Models are pre-trained on IIT-CDIP and fine-tuned on the train set of FUNSD. More implementation details are in the appendix.

\noindent\textbf{Evaluation Metric}. We use F1 score to evaluate models. Exact string matching between our predicted values and the ground-truth ones is used to count true positive, false positive and false negative. If a query has multiple value answers, a prediction is counted as correct if it equals one of them.

\begin{table}[t]
    \centering
    \begin{tabular}{l|c|c|c|c}
         Model& Pretrain & Precision & Recall & F1  \\
         \hline
         \multirow{3}{*}{Baseline}& \small{Bert} & 31.7 & 32.0 & 31.9 \\
         & \small{LayoutLM} & 41.8 & 42.1 & 41.9\\
         & \small{SimpleDLM} & 56.0 & 56.5 & 56.3\\
         \hline
         \multirow{3}{*}{\textbf{Ours}}& \small{Bert} & 35.1 & 35.4 & 35.3\\
         & \small{LayoutLM} & 43.6 & 43.9 & 43.8\\
         & \small{\textbf{SimpleDLM}} & \textbf{60.4} & \textbf{60.9} & \textbf{60.7}\\
         \hline
    \end{tabular}
    \caption{Comparisons on FUNSD when our model and baseline use different pre-trained models.}
    \label{tab:pretrain}
    \vspace{-4mm}
\end{table}

\begin{table}[t]
    \centering
    \begin{tabular}{l|c|c|c|c}
         Model & Query & Precision & Recall & F1  \\
        \hline
        Baseline &\multirow{2}{*}{\small{Exact Key}} &  33.5 & 31.5 & 32.5\\
         \textbf{Ours}& & \textbf{50.5} & \textbf{47.6} & \textbf{49.0}\\
         \hline
        Baseline &\multirow{2}{*}{\small{Field Name}} &  6.0 & 5.6 & 5.7\\
         \textbf{Ours} & & \textbf{21.2} & \textbf{19.9} & \textbf{20.5}\\
         \hline
    \end{tabular}
    \caption{Comparison in the transfer setting. Models are trained on FUNSD and evaluated on INV-CDIP. SimpleDLM is used for both methods.}
    \label{tab:transfer}
    \vspace{-5mm}
\end{table}

\noindent\textbf{Experimental Results}.
The comparisons between our method and the baseline when pre-trained using different approaches are shown in Table~\ref{tab:pretrain}. The performance of both methods are improved largely with SimpleDLM. For the baseline, the F1 score is improved by 14.4\% when replacing the LayoutLM with our SimpleDLM as the pre-trained model. Similarly, using SimpleDLM increases the F1 score of our method by 16.9\% compared to using LayoutLM. Our method outperforms the baseline by 3-4\% using different pre-trained models.

Transferring ability to another dataset is important in real-world applications. We measure this ability by directly evaluating the trained models using FUNSD to the test set of INV-CDIP in Table~\ref{tab:transfer}. As shown, when transferring to a new dataset, the performance of both our method and the baseline drops, compared to the numbers in Table~\ref{tab:pretrain}. When using the exact key as the query, our method largely surpasses the baseline by 16.5\% in F1 score. 

In practice, we may not assume the input queries match exactly with the actual keys shown in a form. Here, we experiment using the field names directly as the queries. Using field names is a more convenient way, since users don't need to design different queries that match keys for different forms. However, field names are more abstract, which makes the problem more challenging. When using the abstract field name as a query, our method achieves 20.5\% F1 which is 14.8\% better than our baseline.

\section{Conclusion}
We introduce a framework for value retrieval with arbitrary queries for form-like documents. It takes a query and the detected OCR words from a document as inputs, models their interactions and predicts the best value corresponding to the input query. We also present SimpleDLM as a pre-training strategy to boost performance. Experimental results show that our method significantly outperforms previous designs in different settings.

\section{Broader Impacts}
This work is introduced to automate the information extraction from forms to improve document processing efficiency. It has positive impacts such as reducing human labor. However, reducing human labor may also cause negative consequences such as job loss or displacement, particularly amongst low-skilled labor who may be most in need of gainful employment. The negative impact is not specific to this work, but should be addressed broadly in the field of AI research.

\bibliography{anthology,custom}

\clearpage
\appendix
\section{Appendix}
\label{sec:appendix}
\renewcommand{\thefigure}{A\arabic{figure}}
\setcounter{figure}{0}
\subsection{Implementation Details}
Our code is implemented using Pytorch. We used Tesseract\footnote{https://github.com/tesseract-ocr/tesseract (Apache License 2.0)} to extract OCR words from documents for IIT-CDIP and INV-CDIP. Since FUNSD provides an official OCR annotation, we use it directly. The total number of query words $M$ and the OCR words are different for different queries and documents. We keep $M+N=512$ and pad with 0s when needed. We follow LayoutLM-base~\footnote{https://github.com/microsoft/unilm/tree/master/layoutlm/deprecated (MIT License)} to setup the structure of our transformers. The fully connected layer used for feature projection has 768 units. Each document is rescaled to [1000, 1000] and the dummy location, $b_0$, is set to [0, 0, 1000, 1000]. Adam is used as the optimizer. During pre-training, the learning rate is 5$e^{-5}$. During fine-tuning, the learning rate is set to 3$e^{-5}$ with weight decay equals to 0.9. SimpleDLM is initialized by LayoutLM and pre-trained on IIT-CDIP using 8 Nvidia A100 GPUs with a batch size of 36 for one epoch. We use a single A100 GPU for fine-tuning, where the training batch size is 8 and the total number of epochs is 45. In our experiments, all the pre-trained models including Bert, LayoutLM and SimpleDLM are base models. The pre-training and fine-tuning of our method take about 10 hours and 2 hours, respectively.

To perform a fair comparison, our method and baseline adopt the same experimental settings except for the interaction strategy. The baseline has the same transformer architecture and the feature projection layer as our method. The transformer takes OCR words with locations as inputs and obtain $\phi_{w_{j}}$ for each word $w_{j}$. And then, the query-value pairing score $\mathbf{s}_j$ is obtained by measuring the distance between the query representation, $\Phi$, and $\phi_{w_{j}}$ as in Equation~\ref{eq:similarity_score}. The query representation is obtained by average pooling over word embeddings extracted from a pre-trained Bert model.

\end{document}